\documentclass[11pt]{article}
\usepackage{fontawesome5}
\usepackage{hyperref}
\usepackage[final]{acl}
\usepackage{booktabs,tabularx}
\usepackage{times}
\usepackage{latexsym}

\usepackage[T1]{fontenc}
\usepackage[utf8]{inputenc}

\usepackage{float} % in preamble
\usepackage{graphicx}
\usepackage{hyperref}

\newcommand{\hficon}{\raisebox{-0.8ex}{\includegraphics[height=2.5ex]{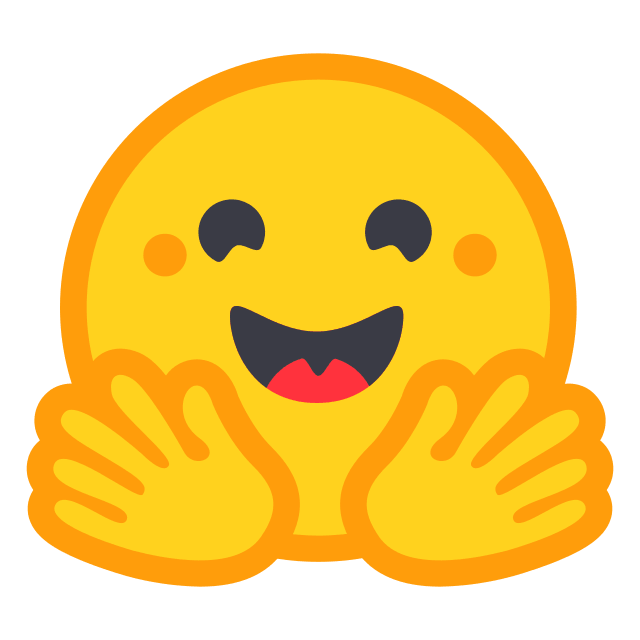}}}
\newcommand{\kaggleicon}{\raisebox{-0.45ex}{\includegraphics[height=2.5ex]{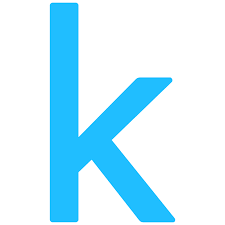}}}

\usepackage{microtype}
\usepackage{xspace}
\usepackage{amsfonts}
\usepackage{amsthm}
\usepackage{amsmath}
\usepackage{amssymb}
\usepackage{cleveref}
\usepackage{bbm}
\usepackage{caption}
\usepackage{subcaption}
\usepackage{xcolor}
\usepackage{tikz}
\usetikzlibrary{bayesnet}
\usepackage{thm-restate}
\usepackage{booktabs}
\usepackage[most]{tcolorbox}
\usepackage{enumitem}
\usepackage{wasysym}
\usepackage{multirow}
\usepackage{scalerel}
\usepackage{titlesec}

\titlespacing*{\section}{0pt}{1ex}{0.5ex}
\titlespacing*{\subsection}{0pt}{0.8ex}{0.4ex}

\usepackage[textsize=tiny]{todonotes}

\usepackage{inconsolata}

\usepackage{graphicx}
\usepackage{hyperref}
\usepackage{cleveref}

\crefname{appendix}{Appendix}{Appendices}
\Crefname{appendix}{Appendix}{Appendices}
\crefname{table}{Table}{Tables}
\Crefname{table}{Table}{Tables}
\title{FMI\_SU\_Yotkova\_Kastreva at SemEval-2026 Task 13:\\ Lightweight Detection of LLM-Generated Code via Stylometric Signals}

\author{Elitsa Yotkova$^1$\thanks{Equal contribution.},
  Violeta Kastreva$^1$\footnotemark[1], Dimitar Dimitrov$^1$, Ivan Koychev$^1$, Preslav Nakov$^2$ \\
  $^1$Faculty of Mathematics and Informatics, Sofia University "St. Kliment Ohridski", Bulgaria \\
  $^2$ Mohamed bin Zayed University of Artificial Intelligence, UAE\\
  \texttt{eyotkova@g.fmi.uni-sofia.bg}, \texttt{vkastreva@uni-sofia.bg}, \\ \texttt{\{ilijanovd, koychev\}@fmi.uni-sofia.bg}, \texttt{preslav.nakov@mbzuai.ac.ae}}

\usepackage{caption}
\begin{document}
\maketitle

% tree-sitter parsing
\begin{abstract}
 % Our final system focuses on capturing descriptiveness signals that frequently appear in LLM-generated code. 

SemEval-2026 Task 13 investigates machine-generated code detection across multiple programming languages and application scenarios, asking participating systems to generalize to unseen languages and domains. This paper describes our participation in Subtask A (binary classification) and explores both pretrained code encoders and lightweight feature-based methods.
We design ratio-based features that are less sensitive to snippet length. To support the extraction of descriptiveness-related signals, we use parsing engines and a programming-language classifier. Additionally, we train a separate code-vs-text line classifier to identify raw natural language segments embedded within samples. We combine a shallow decision tree with heuristic rules derived from data analysis to produce the final predictions. Our approach is computationally efficient, requires only CPU resources for training, and achieves near-instant inference time, offering a lightweight alternative to large pretrained models.
Our code is available at \url{https://github.com/violeta-kastreva/SemEvalTask13-SubtaskA}.
\end{abstract}

\section{Introduction}
The widespread adoption of large language models (LLMs) for code generation has reshaped modern software development. Current models can generate syntactically correct and semantically meaningful programs across many programming languages and domains. This has led to increased presence of machine-generated code in cases such as automated unit test generation \cite{jain2025testgeneval}, code infilling and completion \cite{bavarian2022efficient_fim}, and production development workflows \citep{ dunay2024multiline, frommgen2024resolving}. 
While LLM-based tools offer clear productivity benefits, their ability to author and refine code raises concerns in domains where human authorship is critical. 

In academia, detecting code authorship in programming assignments is increasingly important, as reliance on LLMs undermines educational integrity \cite{sullivan2023chatgpt_higher_ed}. 
Similarly, fair technical hiring requires verifying that submitted artifacts are genuinely human-written. These use cases motivate the need for robust detection of machine-generated code.

In an industrial setting, machine-generated code can pose technical risks when deployed without sufficient oversight. Prior work has shown that LLM-generated code may contain insecure logic, hidden backdoors, or injection vulnerabilities, threatening software reliability and data security
\citep{bukhari2024ai_code_detection, pearce2025_asleep_at_the_keyboard}. 

Distinguishing machine-generated code from human-written one is therefore a critical challenge. Unlike natural language, source code is tightly constrained by syntax and semantics, which can obscure stylistic differences between human and machine authors, which are commonly used for machine-generated text detection. In practice, code detection systems must generalize across programming languages, application domains, and generator families, and many existing approaches degrade substantially in such conditions \cite{orel-etal-2025-droid}.

Recent work has largely focused on pretrained code encoders and large neural models for authorship detection. These models can be effective in in-domain settings, but their performance and reliability can vary under distribution shift, particularly for unseen languages or domains. Moreover, their predictions are often difficult to interpret, making it harder to identify the properties of code that distinguish human-written from machine-generated samples. 

This paper presents our approach to SemEval-2026 Task 13 Subtask A  \cite{orel-etal-2026-semeval-2026}, which targets binary classification of machine-generated versus human-written code under diverse and challenging generalization settings. 

Instead of relying solely on high-capacity neural representations, we explore lightweight, interpretable signals that capture descriptiveness and stylistic leakage commonly observed in LLM-generated code. Our approach is motivated by the observation that machine-generated code often includes verbose comments, explanatory text, or natural-language artifacts that differ in both frequency and structure from human-written code.

%These include the ratio of comment lines to total lines, the proportion of text-like lines to code lines within a snippet, and the relative frequency of verbs in extracted comments
We propose a feature-based detection pipeline centered on ratio-based descriptors that are less sensitive to code length and formatting variability. To ensure robust feature extraction across incomplete or noisy code samples, we employ parsing engines and introduce a dedicated programming language identification classifier. In addition, we train an auxiliary code-versus-text line classifier to detect natural language segments embedded within samples.

Our final system combines shallow machine learning classifiers with heuristic rules, prioritizing robustness and interpretability over model complexity. Through extensive experimentation, we compare pretrained encoder-based representations with feature-driven methods and analyze the strengths and limitations of each approach under cross-language and cross-domain evaluation. On the official leaderboard, our approach is ranked within the top 15\% of submissions.
Our contributions are as follows: 

\begin{itemize}
\item We analyze the limitations of pretrained code encoders for code detection under distribution shift. 
% original text: We introduce ratio-based, interpretable features that capture descriptiveness and stylistic signals characteristic of LLM-generated code.
\item We introduce interpretable ratio-based features that capture descriptiveness and stylistic signals in LLM-generated code, and use them to train classical machine learning classifiers. 
% original text: We design auxiliary classifiers, one for programming language identification and one for code-vs-text detection to support reliable feature extraction across heterogeneous samples.
%\item We design two auxiliary classifiers: one for programming-language identification and one for code-versus-text line detection. %, to enable reliable feature extraction across heterogeneous samples.

\item We construct a dataset for code-vs-text line classification,\footnote{\kaggleicon \ \href{https://www.kaggle.com/datasets/violetakastreva/line-level-code-vs-text-classification-dataset/data}{violetakastreva/line-level-code-vs-text-classification-dataset}} \footnote{\hficon { }\href{https://huggingface.co/datasets/violetakastreva/line-level-code-vs-text-classification}{ violetakastreva/line-level-code-vs-text-classification}}
 enabling line-level discrimination between natural language text and source code. 

\item We present a lightweight detection pipeline that performs competitively in SemEval-2026 Task~13 while remaining simple and interpretable. 
\end{itemize}

\section{Related Work}
Prior work on detecting machine-generated code has emphasized the role of code stylometric features, showing that LLM-generated code differs from human-written code in surface and structural properties such as line statistics, comment usage, and complexity measures \cite{rahman2025automatic_llm_code_detection}. Studies on Claude 3-generated code report that features related to the number of lines, blank lines, comment ratio, and cyclomatic complexity are particularly informative, with generated code often exhibiting simpler structure and distinct comment-to-code patterns compared to human-written code \cite{rahman2025automatic_llm_code_detection}.

Other approaches move beyond stylometry by leveraging model-internal signals. Some methods compute token-level likelihoods from large language models and frame detection as a classification problem over probability maps using vision architectures \cite{xu2025codevision}. In addition, pretrained code encoders are commonly used as feature extractors for supervised classification of code authorship, providing strong semantic representations at the cost of reduced interpretability \citep{rudin2019stop, NGUYEN2024112059}.

\section{Dataset}
We use the dataset provided for SemEval-2026 Task 13 \citep{orel-etal-2025-codet, orel-etal-2025-droid} Subtask A, which targets binary classification between machine-generated and human-written code. Each sample is labeled as either fully human-written or fully machine-generated. The training data consists of algorithmic code snippets written in C++, Python, and Java, while evaluation additionally includes unseen programming languages (Go, PHP, C\#, C, and JavaScript) and unseen application domains (research and production code).

The dataset contains 500K training samples (238K human-written and 262K machine-generated), of which approximately 457K are written in Python, a 100K validation set, and a 500K test set. The participating systems are evaluated using macro-F1 scores across multiple settings that assess generalization across languages and domains.

\section{Methodology}
\begin{figure}[H]
    \centering
    \includegraphics[width=\columnwidth]{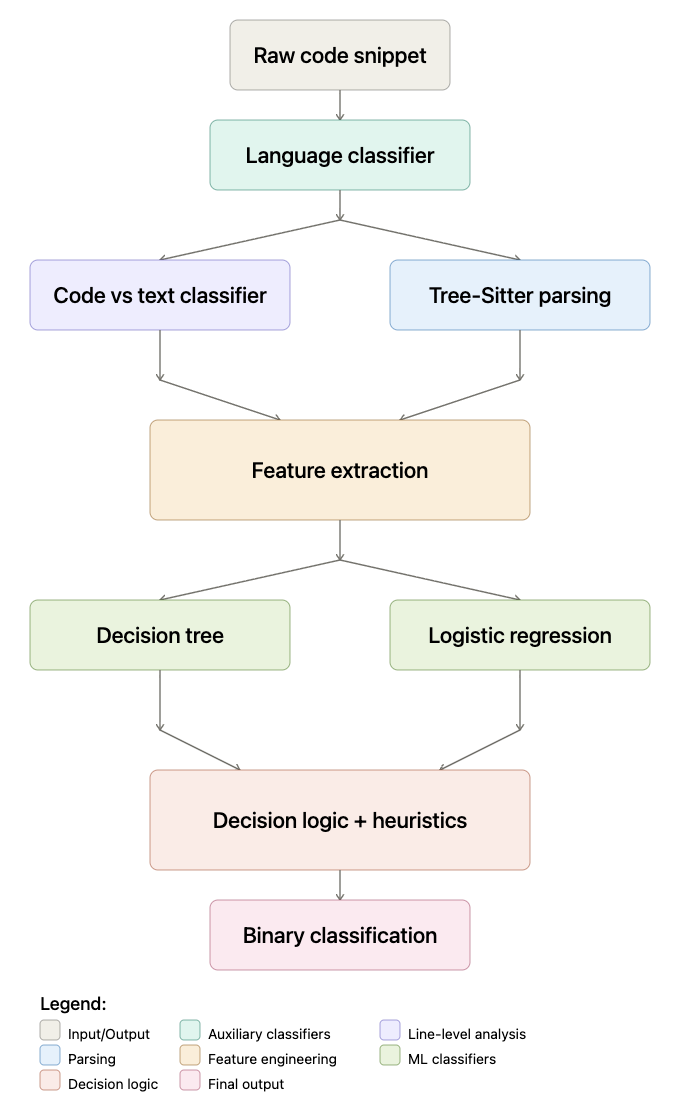}
    \caption{Overview of the proposed machine-generated code detection pipeline.}
    \label{fig:pipeline}
\end{figure}

Our system is built around feature-based representations supported by several auxiliary classifiers to enable robust detection of machine-generated code across multiple programming languages and heterogeneous input formats as shown on \Cref{fig:pipeline}.

%The system first identifies the programming language, extracts stylometric and structural features, applies auxiliary code-vs-text classification, and produces final predictions using shallow classifiers and heuristics.
% IK. Our system is built around feature engineering and several auxiliary classifiers to enable robust detection of machine-generated code across multiple programming languages and heterogeneous input formats.

% In addition, we explore pretrained code embeddings as alternative representations. 

% \dimitar{Can you create a figure with the whole pipeline and how things are built before the final classification - be creative, you can embed actual example in the pipeline or just present it as different parts of a complete architecture \response{Violeta}{Should the figure be present in the appendix or the main text?}}

\subsection{Feature Extraction and Engineering}

We design ratio-based, interpretable, length-normalized features that capture stylistic descriptiveness and structural cues often present in LLM-generated code. \Cref{tab:code_features_1} lists our stylometric ratio features, while \Cref{tab:code_features_2} summarizes our syntactic and structural features, which we derive from parsing and complexity estimates.
% These include the ratio of comment lines to total lines, the proportion of text-like lines relative to code lines, and a verb-to-word ratio computed over extracted comments. Linguistic features are computed using spaCy for tokenization and verb identification. 

\begin{table}[h]
\centering
\setlength{\tabcolsep}{3pt}
\renewcommand{\arraystretch}{1.05}
\footnotesize
\begin{tabularx}{\columnwidth}{lX}
\toprule
\textbf{Feature} & \textbf{Definition} \\
\midrule

% \multicolumn{2}{@{}l}{\textit{Code Stylometry Features}} \\

Comment Ratio &
comment lines to total lines in the code snippet \\

Verb Comment Ratio &
verb tokens to all word tokens in comment text \\

Text-like Ratio &
lines classified as natural language text to total number of code lines \\

Identifier Verb Ratio &
verb tokens to total number of identifier tokens; the identifiers are decomposed into their constituent word units before token counting \\

Comment Code-like Ratio &
comment lines exhibiting code-like patterns to all comment lines \\
\bottomrule

\end{tabularx}
\caption{Code stylometric ratio features computed at the snippet level.}
\label{tab:code_features_1}

\end{table}

\begin{table}[h]
\centering

\setlength{\tabcolsep}{3pt}
\renewcommand{\arraystretch}{1.05}
\footnotesize
\begin{tabularx}{\columnwidth}{lX}
\toprule
\textbf{Feature} & \textbf{Definition} \\
\midrule

% \multicolumn{2}{@{}l}{\textit{Syntactic and Structural Features}} \\

Completeness Score &
an indicator of whether the code snippet forms a syntactically complete unit \\

Error Nodes Near EOF &
concrete syntax tree error nodes occurring near the end of the file to all error nodes \\

Cyclomatic Complexity (Mean) &
average cyclomatic complexity computed across all functions or methods in the snippet \\

% Cyclomatic Complexity (Std) &
% Standard deviation of cyclomatic complexity across all functions or methods in the snippet. \\
% \midrule

Statements-to-LOC &
ratio of the number of statements to the total number of lines of code \\
\bottomrule

\end{tabularx}
\caption{Snippet-level syntactic and structural features derived from parsing and complexity estimates.}
\label{tab:code_features_2}

\end{table}

We extract our linguistic features using spaCy\footnote{\ \href{https://github.com/explosion/spaCy}{https://github.com/explosion/spaCy}} for tokenization and verb identification. For reliable comment extraction, we use the incremental parser Tree-Sitter,\footnote{\ \href{https://tree-sitter.github.io/tree-sitter/}{https://tree-sitter.github.io/tree-sitter/}} which enables parsing across all seven languages in the shared task and remains robust to incomplete or syntactically invalid code snippets. Since Tree-Sitter requires the programming language to be known in advance, we introduce a dedicated language identification component, as described in the next subsection.
% We selected the final feature subset using model-based interpretability: SHAP feature importance for tree-based classifiers and coefficient inspection for logistic regression. The selected features were then fixed\dimitar{What does "fixed" mean in this context? Rephrase accordingly - Features were selected based on ... and used for all experiments...} and used for all experiments reported in this paper.

\subsection{Programming Language Identification}
To support language-specific parsing, we train a programming language classifier using the Rosetta Code dataset \cite{rubin2014rosetta_code_dataset}. It is based on character (3,8) $n$-gram TF.IDF features and a Multinomial Na\"{i}ve Bayes model with a smoothing hyperparameter of 0.1, enabling fast language prediction across all languages included in the task.

\subsection{Code vs. Text Classification}
During our analysis of the data, we observed that some samples contain substantial amounts of raw natural language, mixed with or replacing code. To explicitly capture this behavior, we train a binary code-vs-text line-level classifier that operates at the granularity of individual lines.
For this task, we construct a custom dataset combining code snippets extracted from Stack Overflow posts \cite{fumery2021_stackoverflow_filtered} and natural language samples from Twitch chat data \cite{mowglii2020_twitch_chat_test_data}. Stack Overflow provides diverse, real-world code fragments, while Twitch chat provides informal, conversational text, allowing the classifier to learn robust distinctions between code-like and text-like content.

% This classifier achieves a macro F1-score of 0.96 
We further use a TF.IDF vectorizer with character (3,5)-gram features, combined with a linear classifier, to distinguish code from text. This classifier is computationally lightweight, making it well-suited as an auxiliary component in the pipeline. 

% Due to its speed and strong performance, we favor this approach over embedding-based alternatives, which would introduce additional computational overhead without clear benefits for this subtask. The classifier's predictions are used as signals during feature extraction and decision-making.

\subsection{Classification and Decision Logic}
%For feature-based representations, we train shallow machine learning models, including logistic regression and decision trees. Final predictions are obtained by combining classifier outputs with simple heuristic rules derived from data analysis, such as identifying language-name-only first lines and thresholding text-like content ratios. This design prioritizes interpretability and robustness while remaining compatible with noisy, incomplete, and multilingual code samples. 

We train shallow classifiers, including logistic regression and decision trees. The feature set comprises comment density and the verb-comment ratio within comments, capturing descriptive patterns frequently observed in LLM-generated code.

The final predictions are produced by combining classifier outputs with lightweight heuristics derived from exploratory data analysis. After manual inspection of the training data, we observed a recurring stylistic artifact in the LLM-generated code: the programming language is often emitted as a standalone line preceding the snippet (e.g.,  \verb|python| or  \verb|java|). This pattern appeared in approximately 5k out of the 500k examples. As a sanity check, we first submitted a trivial baseline that predicted only the human-written class (all zeros). We then incorporated a simple heuristic that flags instances with such standalone language markers, yielding an absolute macro-F1 improvement of 0.05 over the baseline. We attribute this to LLMs mimicking Markdown formatting.
% , where code is often presented in fenced code blocks annotated with a language tag (e.g., \verb|``python|), and the tag sometimes appears as a standalone line in the raw snippet.
\begin{table}[h]
\centering
\begin{tabularx}{\columnwidth}{>{\raggedright\arraybackslash}Xc}
\toprule
\textbf{Model} & \textbf{Leak Rate (\%)} \\
\midrule
Qwen2.5-Coder-1.5B-Instruct & 87.28 \\
Phi-3-medium-4k-instruct & 35.78 \\
Phi-3.5-mini-instruct & 59.81 \\
Yi-Coder-1.5B-Chat & 74.39 \\
Qwen2.5-Coder-7B-Instruct & 66.18 \\
\bottomrule
\end{tabularx}
\caption{Leak rate denotes the percentage of all snippets from that model Markdown formatting.}
\label{tab:markdown_leakage_top5}
\end{table}

% Upon further analysis of the training data, we observed that a substantial proportion of examples labeled as LLM-generated contain characteristic lexical markers such as  \emph{Explanation},  \emph{code here}, and  \emph{without comments}, which occur far less frequently in human-labeled instances. 

% Incorporating these cues alongside the standalone programming-language-line heuristic further improved performance. In particular, augmenting our main classifier with these lightweight heuristics further increased macro-F1.

% In particular, we exploit the observation that LLM-generated code frequently begins with the programming language name as a standalone first line (e.g., “\emph{python}”, “\emph{java}”), a pattern rarely observed in human-written code.

% We further design a heuristic based on the text-like content ratio. From the training data, we compute the mean text-like ratio over samples with non-zero values and define an optimistic threshold strictly above this mean. Test samples whose text-like ratio exceeds this threshold are directly classified as LLM-generated, a rule that achieves high precision on held-out validation data.
We also design a heuristic based on the text-like content ratio. On the training set, we compute the mean text-like ratio over samples with non-zero values. The mean text-like ratio is 0.07 for human-written labeled examples and 0.17 for LLM-generated code (considering only non-zero text-like ratio values). Based on these statistics, we define an optimistic threshold strictly above both means and set it to 0.3.
%We validate this threshold via distributional analysis. A Kolmogorov-Smirnov test confirms the text-like ratio distributions differ significantly between classes (D=0.40, p<0.001). While Youden's J optimization yields 0.008, we use 0.3 to prioritize precision: this achieves 99.9\% specificity, flagging only samples at the 90th percentile of LLM-generated code's text-like content.
A Kolmogorov-Smirnov test confirms that the text-like ratio distributions differ significantly between classes ($D=0.40$, $p<0.001$). While Youden's J optimization yields a threshold of 0.008, we use 0.3 to prioritize specificity. This threshold achieves 99.9\% specificity on human-written code and flags only the top 10.3\% of LLM-generated samples with the highest text-like ratios.
\begin{figure}[H]
    \centering
    \includegraphics[width=\columnwidth]{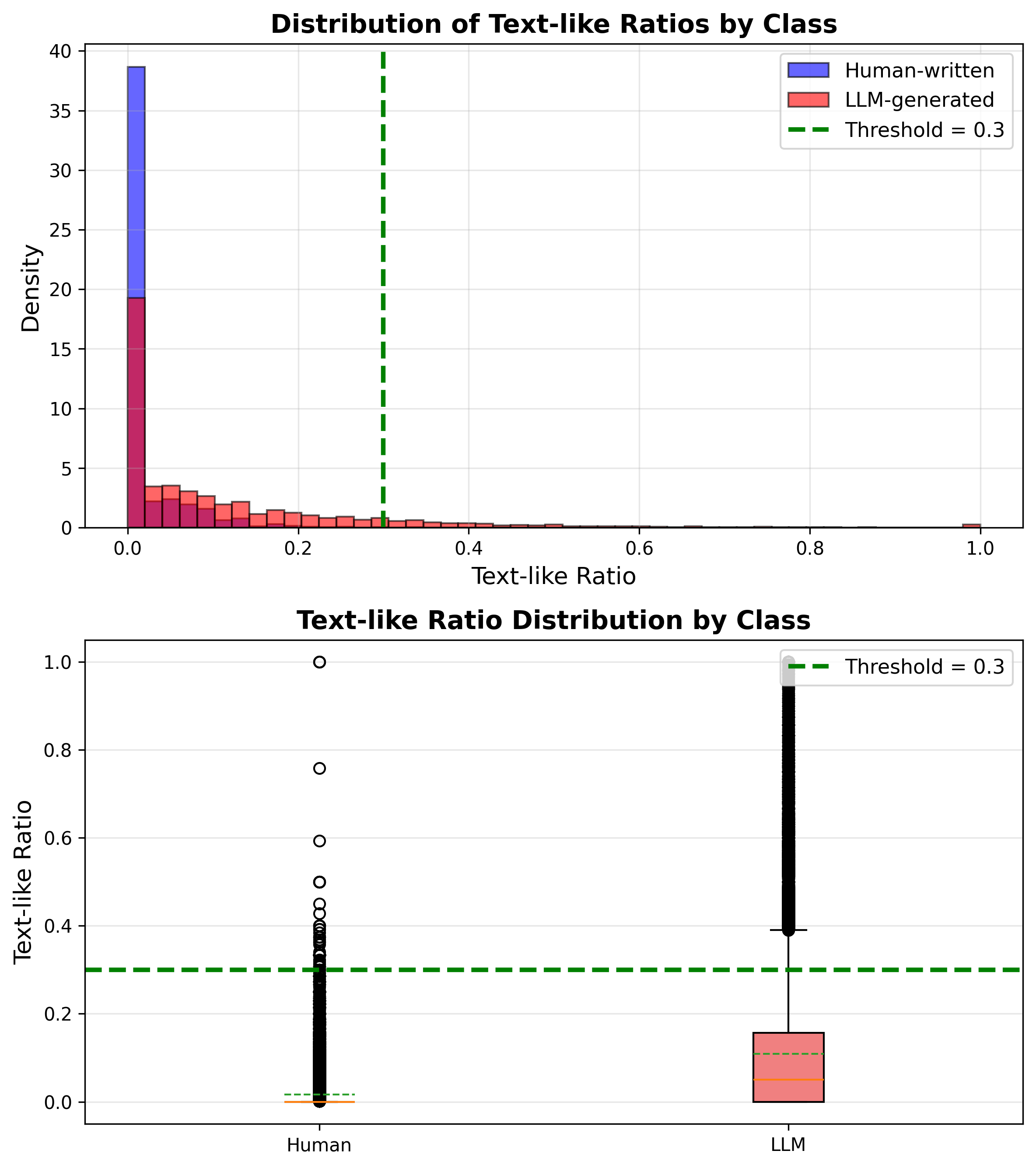}
    \caption{Text-like ratio distributions by class. (a) Histograms and (b) box plots show significant differences (K-S test: D=0.40, p<0.001). Threshold = 0.3 (green line).}
    \label{fig:text_like_ratio}
\end{figure}

During inference, test samples whose text-like ratio exceeds 0.3 are directly classified as LLM-generated. As illustrated in \Cref{fig:pipeline}, these heuristics are applied alongside classifier predictions to produce the final binary classification.

% This rule achieves high precision on held-out validation data.

To account for systematic differences across snippet lengths, we partition the code samples into small, medium, and large buckets in terms of line count and apply separate feature scaling to each bucket. The logistic regression decision threshold is tuned on the validation set to optimize the precision–recall trade-off, whereas the decision tree classifier is constrained to depth 2 and trained using the Gini impurity criterion in order to preserve the interpretability.

\section{Experiments}

% In our experiments, we evaluate pretrained code embeddings, handcrafted features, and auxiliary signals for machine-generated code detection, using the official SemEval-2026 Task 13 train/validation split and evaluation protocol. All experiments were conducted on Google Colab using only the CPU, with the complete training pipeline taking approximately 2 hours and inference remaining near-instant. \dimitar{Here having a visual presentation of the pipeline would help people understand if 2 hours is a lot or not. If people did not read carefully, 2 hours could seem a lot but what does 2 hours actually include is important}

We evaluate the utility of using pretrained code embeddings, handcrafted features, and auxiliary signals for machine-generated code detection on the official SemEval-2026 Task 13 train/validation split and following the official evaluation protocol. The full pipeline, including embedding extraction, runs in approximately 2 hours on a CPU; however, computing the embeddings with large pretrained models can incur substantially higher computational cost, even on a GPU.

\subsection{Pretrained Code Embeddings}

%These embeddings are considered independently and are not integrated into the final feature-based pipeline. 
% We experiment with pretrained code encoders as standalone representations for machine-generated code detection. Specifically, we extract embeddings from GraphCodeBERT \cite{guo2021graphcodebert}, CodeT5+ \cite{wang2023codet5plus}, and nomicai/CodeRankLLM \cite{suresh2025cornstackhighqualitycontrastivedata}, and evaluate them using standard supervised classifiers. Although these models achieve strong training performance, they exhibit limited generalization and are therefore not integrated into the final feature-based pipeline. Detailed quantitative results are provided in Appendix~\ref{sec:appendix_additional_results} (\Cref{tab:appendix_model_results})

We evaluated the pretrained code encoders as standalone representations for machine-generated code detection, extracting embeddings from GraphCodeBERT \cite{guo2021graphcodebert}, CodeT5+ \cite{wang2023codet5plus}, and CodeRankLLM \cite{suresh2025cornstack} with standard classifiers. \cref{tab:models_results} reports results on the initial 1,000-example development test subset. We also combined encoder embeddings with our feature-based pipeline, but this yielded no improvement. Analysis of Decision Tree feature importances and Logistic Regression coefficients showed that the embedding features contributed negligibly compared with our handcrafted features. Although these pretrained models performed well on the training data, they generalized poorly. We attribute this largely to language shift: 91.4\% of the training set consists of Python snippets, whereas the official test set spans seven programming languages. As a result, models may overfit to language-specific cues that transfer weakly across languages. For instance, indentation patterns may be a strong signal for distinguishing LLM-generated from human-written code in Python, where formatting is syntactically meaningful, but such cues are far less informative in languages such as C\#, JavaScript, or C++.
%Despite strong training performance, they generalize poorly and are not included in the final feature-based pipeline. 
% Detailed results appear in \cref{appendix:appendix_additional_results} (\Cref{tab:appendix_model_results}).

\begin{table}[tbh]
\centering
\footnotesize
\begin{tabularx}{\columnwidth}{>{\raggedright\arraybackslash}Xcc}
\toprule
\textbf{Model} & \textbf{Validation} & \textbf{Test} \\
\midrule

GraphCodeBERT & 98.21 & 25.11
 \\
CodeT5+ encoder & 97.46 & 49.39 \\
nomicai/CodeRankLLM & 95.27 & 33.65 \\
% CodeT5+ encoder + comment-ratio & X & X \\
% \midrule
% GraphCodeBERT + comment-ratio & X & X \\
\bottomrule
\end{tabularx}
\caption{Pretrained code embeddings: best downstream classifier results (validation vs test macro-F1).}
\label{tab:models_results}

\end{table}

% \dimitar{Do not dedicate a whole subsection and just say - this didnt work. Present all results in a table, later in the result discussion/analysis talk about why these were not included. You are jumping straight to conclusions from experiments without any results presented! \response{eli}{@Violeta add table showing the experiments with the pretrained models} }

\subsection{Feature Selection}

% We select the final feature subset based on model-based interpretability, relying on SHAP feature importance for tree-based classifiers and coefficient inspection for logistic regression. Based on this analysis, we retain the two best-performing features for the final system: the comment ratio and the verb comment ratio. We additionally consider several auxiliary features that provide complementary signals in intermediate models, which we describe in detail below.

We select the final feature subset based on model-based interpretability, relying on SHAP feature importance for tree-based classifiers and coefficient inspection for logistic regression. The initial feature pool comprised approximately 30 stylometric and syntactic features, including measures such as nesting-mean, return-statement-ratio, identifier-entropy, and related statistics. We also examined pairwise feature correlations to identify redundant signals and reduce multicollinearity. Although this broader set captured diverse properties of the source code, only a small subset was consistently informative for the final classification task. Based on this analysis, we retain the two best-performing features for the final system: the comment-ratio and the verb-comment-ratio. We additionally consider several auxiliary features that provided complementary signals in intermediate models, which we describe in detail below.

\paragraph{Errors near EOF ratio}
We introduce a syntax-based feature that captures the distribution of parsing errors near the end of a code snippet. We collect all nodes labeled as parsing errors and compute the proportion of those whose spans overlap with the final 20\% of the file. This ratio is designed to reflect incomplete or abruptly terminated code, a pattern frequently observed in LLM-generated snippets. Incorporating this feature yields stable performance across splits, with results nearly identical to our best-performing configuration.
% For example, languages such as PHP and JavaScript yield very few or no error nodes under the same conditions, reducing the reliability of this signal across languages and contributing to reduced generalization.

\paragraph{Completeness score}

To capture incomplete or abruptly terminated code without relying on Tree-Sitter parsers, we prompt Qwen2.5-7B-Instruct-1M \cite{yang2025qwen25_1m} to assess the completeness of the final line in a code snippet. The model assigns a score of 0 to complete lines, 0.5 to uncertain cases, and 1 to incomplete lines.
% , which we then used as an additional feature.
While this approach successfully captures some forms of truncated code, it is computationally expensive and does not really yield improvements on the test set.

\paragraph{Computing cyclomatic complexity}

We compute the cyclomatic complexity by heuristically extracting function or method bodies from signatures and block structure using language-agnostic patterns. For each function, the complexity is defined as one plus the number of decision points (e.g., conditionals, loops, logical operators, and exception handling), computed only over lines that are classified as code, with comments and string literals removed. We aggregate the function-level values within each snippet and use the means as global features. These statistics capture structural differences but provide only a moderate signal, particularly for production code where complexity varies widely across styles and domains.

\begin{table}[h]
\begin{tabularx}{\columnwidth}{>{\raggedright\arraybackslash}Xcc}
\toprule
\textbf{Feature} & \textbf{Validation} & \textbf{Test} \\
\midrule

Error near EOF ratio & 64.75 & 63.90 \\

Completeness score & 64.14 & 63.26 \\

Cyclomatic complexity mean & 53.59 & 63.16 \\
% \midrule

% Cyclomatic complexity mean & X & X \\
% \midrule

% Cyclomatic complexity mean & X & X \\

\bottomrule

\end{tabularx}
\caption{Performance of additional features combined with the main features (Comment Ratio and Verb Comment Ratio) using Logistic Regression.}
\end{table}
\paragraph{Summary}
We observe similar behavior for other structural features, including Identifier Verb Ratio and Statements-to-LOC. Although they capture certain stylistic and syntactic properties, their contribution to the final classification decision is limited. In practice, they either provide only a weak signal or overfit to superficial patterns in the training data, which reduces their robustness across domains.

\subsection{Logistic Regression Threshold Tuning}

As described earlier, we partition the code samples into three buckets by the number of code lines and apply separate feature scaling to each group. For each bucket, we optimize a decision threshold for a logistic regression classifier on the validation set and assign labels at inference time using the bucket-specific threshold. This strategy proves highly effective in improving validation performance. However, the optimal thresholds learned for individual buckets on the validation set do not generalize consistently to the test set, indicating sensitivity to distributional differences across code-length regimes and limiting generalization. 

%and the final competition score

% \subsection{Heuristic based approach}

% After manual inspection of the training data, we observed a recurring stylistic artifact of LLM-generated code: the programming language is often emitted as a standalone line preceding the snippet (e.g., “python” or “java”). This pattern appeared in approximately 5k out of 500k examples. As a sanity check, we first submitted a trivial baseline that predicted only the human-written class (all zeros). We then incorporated a simple heuristic that flags instances containing such standalone language markers, which yielded an absolute macro-F1 improvement of approximately 0.05 over the baseline, highlighting the predictive value of this surface-level signal. We attribute this behavior to LLMs mimicking Markdown formatting, where code is often presented in fenced code blocks annotated with a language tag (e.g., \verb|``python|
% ), and the tag sometimes appears as a standalone line in the raw snippet.

% Upon further analysis of the training data, we observed that a substantial proportion of examples labeled as LLM-generated contain characteristic lexical markers such as  \emph{Explanation},  \emph{code here}, and  \emph{without comments}, which occur far less frequently in human-labeled instances. Incorporating these cues alongside the standalone programming-language-line heuristic further improved performance. In particular, augmenting our main classifier with these lightweight heuristics further increased macro-F1.

\section{Results}

Detailed per-model results are provided in \cref{appendix:appendix_additional_results} (\Cref{tab:appendix_model_results}); below, we summarize the main findings and discuss the key performance trends. 
% \dimitar{why is the table in the Appendix. You do not have issues with paper length yet. \response{Violeta}{We already have no space left... but if we have, after modifications, we'll get it back here}}

\subsection{Code-vs-Text Classifier}

The code-vs–text line classifier achieves strong performance on the validation set, with a macro-averaged F1 score of 96.07 and an accuracy of 97.11. On a manually annotated subset of 30 examples drawn from the test set, performance decreases to an accuracy of 89.47 and a macro-averaged F1 score of 86.43.This reduction is largely due to ambiguous cases in the competition’s test data, where certain lines contain repetitive or non-linguistic character sequences (e.g., "\emph{i…i…i…}", "\emph{it's 4am  i've been  there ? what do  i have ?}") originating from code snippets. Although such lines are labeled as text in the dataset, classifying them as code is reasonable and expected. Due to its speed and strong performance, we favor this approach over embedding-based alternatives, which would introduce additional computational overhead without clear benefits for this subtask. 
%The classifier's predictions are used as signals during feature extraction and decision-making.
%, reflecting the inherent difficulty of line-level code-vs–text discrimination in heterogeneous samples.
%On a manually annotated sample of the test set, performance drops to an accuracy of 89.47 and a macro-average F1 score of 86.43.

\subsection{Language Identification Classifier}
The programming language identification classifier achieves an accuracy of 95.10 on a test set of 1k examples derived from the public test, with labels obtained by prompting the OpenAI ChatGPT-5.2 API. This auxiliary annotation was introduced solely for testing purposes and as a sanity check; model validation was conducted on the Rosetta Code dataset. While most predictions are correct, some misclassifications are critical in principle, as different programming languages may induce distinct error nodes or comment signatures in the Tree-Sitter grammars used for parsing. The most frequent confusion occurs between C and C++, and between C\# and Java. However, these errors have a limited impact on downstream performance, as the affected languages share highly similar grammatical structures in Tree-Sitter, yielding comparable parse trees for feature extraction.

\subsection{Code Classification}
For the binary classification of machine-generated versus human-written code, the decision tree classifier achieves a macro-averaged F1 score of 65.62. In contrast, logistic regression with adjusted decision thresholds attains a macro-averaged F1 score of 63.16. Incorporating the heuristic-based decision function yields an absolute improvement of approximately 2\% in macro-averaged F1. Overall, our system achieves a macro-F1 of 67.35, demonstrating benefits of combining lightweight classifiers with heuristics.

\begin{table}[h]
\centering
\label{tab:code_results}
\setlength{\tabcolsep}{3pt}
\renewcommand{\arraystretch}{1.05}
\footnotesize
\begin{tabularx}{\columnwidth}{>{\raggedright\arraybackslash}Xc}
\toprule
\textbf{Classifier} & \textbf{Macro-F1} \\
\midrule
Logistic Regression, default treshold & 63.16  \\

Logistic Regression, threshold = 0.65 & 64.59
 \\

Decision Tree, depth = 2 & 65.62 \\

Decision Tree, depth = 2 + heuristic &
\textbf{67.35} \\

% \midrule
 % LinearSVC & 0.63265 \\

\bottomrule

\end{tabularx}
\caption{Classifier overview with Comment Ratio and Verb Comment Ratio as features.}
\end{table}

\section{Conclusion and Future Work}
% We presented a lightweight, interpretable system for SemEval-2026 Task~13 Subtask~A. The approach uses ratio-based stylometric features, supported by a programming-language identification model and a line-level code-vs-text classifier. A shallow classifier with a small set of heuristics achieves a macro F1-score of 0.673 on the official test set, while enabling CPU-only, low-latency inference.

We presented a lightweight, interpretable system for SemEval-2026 Task~13 Subtask~A based on ratio-driven stylometric features, supported by language identification and line-level code–text classification. A shallow classifier with minimal heuristics achieves a macro-F1 of 67.35 on the official test set with CPU-only, low-latency inference.

% (machine-generated code detection)
% for Tree-Sitter parsing

In future work, we plan to improve multi-core CPU utilization by parallelizing feature extraction and reducing Python overhead. We further plan to combine these features with LLM-based signals (e.g., selective fallback or distillation) to improve robustness under distribution shift.

\section*{Acknowledgements}

This research is partially funded by the EU NextGenerationEU, through the National Recovery and Resilience Plan of the Republic of Bulgaria, project SUMMIT, No BG-RRP-2.004-0008.

\bibliography{anthology, custom}

\appendix
\onecolumn
\crefalias{section}{appendix} % article class
\section{Main-Task Model Results}

\label{appendix:appendix_additional_results}
\cref{tab:appendix_model_results} reports detailed main-task results on the official test set, comparing macro-F1 across classifiers and training setups. Notably, the strongest performance comes from linear models (and a shallow decision tree), while more flexible nonlinear baselines (random forests, MLP) perform worse---suggesting the stylometric ratio features induce a linear decision boundary and that additional nonlinearity tends to reduce robustness under distribution shift.

\begin{table}[h]
\centering
\setlength{\tabcolsep}{3pt}
\renewcommand{\arraystretch}{1.05}
\footnotesize

\begin{tabularx}{\columnwidth}{lX>{\centering\arraybackslash}p{2.2cm}@{}}
\toprule
\textbf{Model} & \textbf{Setup} & \textbf{Macro-F1} \\
\midrule
Logistic Regression & solver=liblinear, bucket scaling + tuned decision threshold 0.65 & 64.59 \\
Decision Tree & depth=2 & \textbf{65.62} \\
Linear SVC & default parameters & 63.27 \\
Random Forest & 11 trees, depth=2 & 62.25 \\
Logistic Regression & solver=liblinear & 63.16 \\
MLP & (10,5), tanh, adam, alpha=0.0001 & 60.25 \\
\bottomrule
\end{tabularx}
\caption{Results for the main task. For logistic regression, \emph{tuned decision threshold} refers to selecting the probability cutoff on the validation set to maximize macro-F1 (rather than using the default value of 0.5).}
\label{tab:appendix_model_results}
\end{table}

% \begin{table}[h]
% \centering
% \label{tab:models_results}
% \setlength{\tabcolsep}{3pt}
% \renewcommand{\arraystretch}{1.05}
% \footnotesize

% \begin{tabularx}{\columnwidth}{@{}lXX@{}}
% \toprule
% \textbf{Model} & \textbf{Validation} & \textbf{Test} \\
% \midrule

% GraphCodeBERT & 0.98 & 0.25
%  \\
% \midrule

% CodeT5+ encoder & 0.97 & 0.49 \\
% \midrule

% nomicai/CodeRankLLM & 0.95 & 0.33 \\
% % CodeT5+ encoder + comment-ratio & X & X \\
% % \midrule
% % GraphCodeBERT + comment-ratio & X & X \\
% \bottomrule
% \end{tabularx}
% \caption{Pretrained code embeddings: best downstream classifier results (validation vs test macro-F1).}
% \end{table}
\end{document}